\documentclass[10pt,conference]{IEEEtran}
\IEEEoverridecommandlockouts
\usepackage{amsfonts}
\usepackage{setspace}
\usepackage{afterpage}

\usepackage{setspace}
\usepackage{amssymb}
\usepackage{epsfig}
\usepackage{graphicx}
\usepackage{xcolor}
\usepackage{algorithm}
\usepackage{algorithmicx}
\usepackage{algpseudocode}
\usepackage{amsmath}

\floatname{algorithm}{Algorithm}

\usepackage[ps2pdf,
bookmarks=false,
bookmarksnumbered=false, 
bookmarksopen=false, 
colorlinks=false]{}

\begin{document}

\title{Anomaly Detection and Sampling Cost Control via Hierarchical GANs}

\author{Chen Zhong, M. Cenk Gursoy, and Senem Velipasalar
\\Department of Electrical Engineering and Computer Science,
Syracuse University, Syracuse, NY 13244
\\Email: czhong03@syr.edu, mcgursoy@syr.edu, svelipas@syr.edu
\thanks{The information, data, or work presented herein was funded in part by National Science Foundation (NSF) under Grant 1618615, Grant 1739748, Grant 1816732 and by the Advanced Research Projects Agency-Energy (ARPA-E), U.S. Department of Energy, under Award Number DE-AR0000940. The views and opinions of authors expressed herein do not necessarily state or reflect those of the United States Government or any agency thereof.}
}

\maketitle

\begin{abstract}
Anomaly detection incurs certain sampling and sensing costs and therefore it is of great importance to strike a balance between the detection accuracy and these costs. In this work, we study anomaly detection by considering the detection of threshold crossings in a stochastic time series without the knowledge of its statistics. To reduce the sampling cost in this detection process, we propose the use of hierarchical generative adversarial networks (GANs) to perform nonuniform sampling. In order to improve the detection accuracy and reduce the delay in detection, we introduce a buffer zone in the operation of the proposed GAN-based detector. In the experiments, we analyze the performance of the proposed hierarchical GAN detector considering the metrics of detection delay, miss rates, average cost of error, and sampling ratio. We  identify the tradeoffs in the performance as the buffer zone sizes and the number of GAN levels in the hierarchy vary. We also compare the performance with that of a sampling policy that approximately minimizes the sum of average costs of sampling and error given the parameters of the stochastic process. We demonstrate that the proposed GAN-based detector can have significant performance improvements in terms of detection delay and average cost of error with a larger buffer zone but at the cost of increased sampling rates.
\end{abstract}


\section{Introduction}
Anomaly detection has been extensively studied in various fields, with applications in different domains. For instance, the authors in \cite{rajasegarar2008anomaly} provided a survey of anomaly detection techniques in wireless sensor networks. In \cite{kanev2017anomaly}, the problem of anomaly detection in home automation systems is reviewed. Recently, learning-based approaches have been proposed as well. Considering the cybersecurity threats, authors in \cite{puzanov2018deep} proposed deep reinforcement one-shot learning for change point detection to address scenarios in which only a few training instances are available, for example, in the zero-day attack. And a random forest machine learning algorithm is presented in \cite{8666450}  to effectively detect compromised loT devices at distributed fog nodes. Moreover, in \cite{moustafa2019outlier} an adversarial statistical learning method has been proposed to detect slight changes in the statistical parameters caused by the attack data.

In the literature, anomaly detection is also conducted among multiple processes. In such cases, active sequential hypothesis testing problem is often modeled as a partially observable Markov decision process, and the reinforcement learning algorithms are applied to dynamically select the processes to be tested. For instance, in \cite{kartik2018policy} and \cite{9013223}, the application of deep Q-networks and actor-critic deep reinforcement learning have been investigated, respectively.

Furthermore, anomaly detection in multivariate time series has attracted interest recently. In such problems, deep learning algorithms are trained by both normal and abnormal training data, and used as classifiers to detect and diagnose the anomalies \cite{wen2019time}, \cite{zhang2019deep}. Different from the deep learning-based detectors, the GAN-based framework proposed in \cite{8931714} and \cite{190104997} is only trained with the normal dataset, and estimate the probability of the anomaly.


Finally, several recent studies have taken the cost of the detection process into consideration. For instance, in \cite{chen2019active}, the performance measure is the Bayes risk that takes into account not only the sample complexity and detection errors, but also the costs associated with switching across processes. In \cite{8721562}, the cost is expressed as a function of the duration of the anomalous state. Considering the cost of both sampling and errors, authors in \cite{9013908} proposed a deep reinforcement learning-based policy for significant sampling with applications to shortest path routing.

Motivated by these studies, we in this paper propose a novel GAN-based anomaly detector that has a prediction capability and detects threshold crossing in a stochastic time series without requiring the knowledge of its statistics. We evaluate the performance of this detector, determine the incurred sampling and delay costs, identify the key tradeoffs, and compare with existing strategies.

\section{System Model} \label{sec:model}
As noted above, we consider anomaly detection as the detection of crossing a threshold $\Gamma$ in a stochastic time series. Specifically, we assume that an anomaly occurs when the monitored stochastic process exceeds or falls below $\Gamma$. Such anomaly detection is required, for instance, in remote monitoring using sensors in smart home, smart city, e-Health, and industrial control applications. In these cases, the monitored process can be modeled as a stochastic process and it is very important to accurately and timely detect if the process (describing e.g., the patient's health in remote health monitoring or the operational characteristics of the power grid in a smart grid application) crosses a threshold. To react to the changes immediately, the system can continuously monitor the environment. However, this will lead to very high sampling, sensing and also communication costs (e.g., if the sensing results need to be sent to a remote processing center). Alternatively, the system can observe and sample the process intermittently. In this case, the sampling can be nonuniform with the sampling rate depending on how close the values are to the threshold $\Gamma$. With this approach, the sampling/sensing cost will be reduced but there will be a higher risk of delay in the threshold-crossing detection. Therefore, there exists a tradeoff between sampling and delay costs and this should be addressed by taking the delay cost into account when making the sampling decisions.

While our framework is applicable to any process or time series, we consider the Ornstein-Uhlenbeck (OU) process (which has applications in physical sciences, power system dynamics, financial mathematics) in this paper in order to be more concrete in our discussions. Additionally, a sampling policy for the OU process is previously derived in \cite{8719977} under the assumption of complete statistical knowledge, and we will compare the performance of the proposed hierarchical GAN framework with that of this policy. OU process can be expressed as
\begin{equation}
d \mathit{x}(t) = \theta (\mu - \mathit{x}(t)) d t + \sigma d W(t)
\end{equation}
where $\mu$ is the mean of the time series, $\theta$ is the speed of mean reversion that scales the distance between $\mathit{x}(t)$ and the mean $\mu$, and $\sigma$ is the volatility to scale the Wiener process $W(t)$. Specifically, we set the mean $\mu$ as $0$.

The cost of error due to delayed detection can be defined as the area enclosed between the actual crossing point $\mathit{x}(T_{\text{true}})$ and the detected crossing point $\mathit{x}(T_{\text{detect}})$, where $T_{\text{true}}$ is the actual time instant at which the threshold is crossed, and $T_{\text{detect}}$ is the time instant when threshold-crossing is detected. Therefore, when the threshold is $\Gamma = 0$, the cost of error can be computed as
	\begin{equation}\label{eq:cost}
	C = \int_{T_{\text{true}}}^{T_{\text{detect}}} |\mathit{x}(t)| dt.
	\end{equation}
Note that the cost of error due to delayed detection is proportional to the value of the process $x$ and the gap between  $T_{\text{detect}}$ and  $T_{\text{true}}$.

In \cite{8719977}, based on the assumption that the parameters of the OU process are known, a policy is derived to control the sampling time. This policy makes use of the OU process parameters and the current sample to estimate the subsequent sampling time that minimizes the sum of the average costs of error and sampling. Specifically, under certain conditions and assuming $\Gamma = 0$, an approximate solution for the next sampling time is given as
\begin{equation}
T^{*}_{1}(\mathit{x}(t)) = T^* + \frac{1-e^{-\theta T^*}}{\sqrt{1-e^{-2 \theta T^*}}} \frac{\sqrt{\pi}}{\sigma \sqrt{\theta}} \mathit{x}(t) \label{eq:approx_solution}
\end{equation}
where $T^* = (\frac{18 \pi c_s^2}{\sigma^2})^{\frac{1}{3}}$, and $c_s$ is a predefined sampling cost.

\begin{figure}
	\centering
	\includegraphics[width=1.0\linewidth]{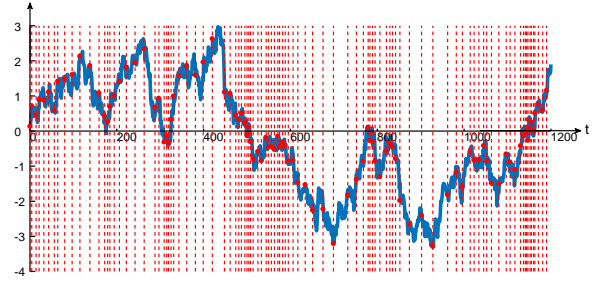}
	\caption{Sampling of an OU process with parameters
		$ \mu = 0$, $ \sigma= 0.5$, $ \theta = 0.025$, and sampling cost $c_s = 0.1$.}
	\label{fig:example}
\end{figure}

In Fig. \ref{fig:example}, we demonstrate such sampling process using (\ref{eq:approx_solution}). The blue curve is the time series $\mathit{x}(t)$, red vertical lines indicate the sampling time instants, and the red dots are the samples collected by the policy. It is obvious that when the value of $\mathit{x}(t)$ approaches the threshold $\Gamma = 0$, the policy samples more frequently, and when $\mathit{x}(t)$ moves far away from the threshold, the policy samples less frequently. Such nonuniform sampling policy provides an effective solution, balancing the detection accuracy and the sampling cost, However, the parameters of the OU process may not be known in practice, and this renders the optimal policy inapplicable. In such cases, data-driven approaches are needed. Considering these scenarios, we in this paper propose a GAN-based framework, that does not require any information on the OU process, to control the sampling time. Indeed, the proposed approach is quite general and applicable to any process. Basically, in this framework, the current sample $\mathit{x}(t)$ will be fed to the GAN, and the samples in the following $N$ time instants will be predicted. And based on the predictions, the next sampling time will be estimated.

We denote the set of predictions obtained at time $t$ as $\{\hat{\mathit{x}}(t+1), \hat{\mathit{x}}(t+2), \dots, \hat{\mathit{x}}(t+N)  \}$, where $N$ denotes prediction length, and $\hat{\mathit{x}}(t+\Delta t)$ is the prediction of $\mathit{x}(t+\Delta t)$ for $\Delta t = 1, 2, \dots, N $. Then, each element in the prediction set will be compared with the threshold $\Gamma$ to see if it is a threshold-crossing point, and to decide the next sampling time. We denote the next sampling time as $T(\mathit{x}(t))$, which can be expressed as
\begin{align}\label{eq:nextTime}
T(\mathit{x}(t))=
\begin{cases}
t + \Delta t, \text{if } \hat{\mathit{x}}(t+\Delta t) \text{ is a crossing point}  \\
\hspace{1.5cm} \text{for some $\Delta t \in [1, 2, \dots, N]$}\\
t + N + 1 , \text{if no crossing point is predicted}
\end{cases}.
\end{align}

Note that the accuracy of such prediction is critical in deciding when to take the new sample. And the mean squared error in the prediction can be expressed as
\begin{equation}\label{eq:sumLoss}
\frac{1}{N} \sum_{\Delta t = 1}^{N} (\hat{\mathit{x}}(t+\Delta t) - \mathit{x}(t+\Delta t) )^2.
\end{equation}

\section{Hierarchical GAN Framework}

\subsection{Preliminaries}\label{subsec: GAN}
We first describe the general structure of a GAN \cite{NIPS2014_5423}. In particular, GAN consists of two neural networks: a generator $G$, that is used to capture the statistical features of the data; and a discriminator $D$ that is used to estimate the probability that a sample comes from the training data rather than the generative model $G$ to evaluate the generative policy.

We first define a sample space $\mathcal{S}$ with a probability distribution $ p(\mathit{s}|\mathit{x})$, where $\mathit{s}$ is a set of samples corresponding to the real data $\mathit{x}$ in the training data set. The generator $G$ maps the sample into the real data space:
\begin{equation}
G(\mathit{s}; \psi) : \mathit{s} \rightarrow \hat{\mathit{x}},
\end{equation}
where $\psi$ denotes the parameters of the generator neural network, and the $\hat{\mathit{x}}$ is a projection (of real data $\mathit{x}$) generated by the generator $G$.

The discriminator $D(\tilde{\mathit{x}};\omega)$ estimates the probability of the input $\tilde{\mathit{x}}$ coming from the real data set, where $\omega$ denotes the parameters of the discriminator neural network, and the input $\tilde{\mathit{x}}$ can be either the real data $\mathit{x}$ or the generated data $\hat{\mathit{x}}$. A good discriminator $D$ is expected to be able to distinguish the generated data from the real data, i.e., the estimated probability should be very small if the input is the generated data and should be close to $1$ if the input is from the real data. Therefore, the discriminator is aimed at minimizing the objective function given as
\begin{equation}
\mathcal{L}_D = - (\log (D(\mathit{x})) + \log (1-D(\hat{\mathit{x}}))).
\end{equation}

On the other hand, for a generator $G$ that has the goal to learn the real data distribution, the evaluation $D(\hat{\mathit{x}})$ acts as a guidance on the update of the generative model. Thus, a good generator $G$ should be able to make the generated data indistinguishable  from the real data to the discriminator $D$. For this purpose, the generator $G$ seeks to maximize $D(\hat{\mathit{x}})$, or equivalently minimize the following objective function:
\begin{equation}\label{eq:l_g}
\mathcal{L}_G =  \log (1-D(\hat{\mathit{x}})).
\end{equation}
The workflow of GAN is presented in Algorithm \ref{alg:GAN} below.

\begin{algorithm}
	\caption{Workflow of GAN}
	\label{alg:GAN}
	\begin{algorithmic}
		
		\State Initialize the generator $G(\mathit{s}; \psi)$ with the parameters $\psi$, and the the discriminator $D(\mathit{x};\omega)$, parameterized by $\omega$.
		
		\For{$T = 1 : \text{Maximum episode}$}
		\State Fetch the sample set $\mathit{s}(t)$ and the corresponding real data set $\mathit{x}(t)$.
		\State Use  the generator $G$ to generate the projection of the real data: $\hat{\mathit{x}}(t) = G(\mathit{s}(t) ; \psi(t))$
		\State Feed the projection $\hat{\mathit{x}}(t)$ and the real data $\mathit{x}(t)$ to the discriminator $D$, and obtain the estimated probabilities of both data being real data.
		\State Update the discriminator by descending the stochastic gradient:
		\State $- \nabla_{\omega} (\log (D(\mathit{x}(t) ; \omega(t)) + \log (1-D(\hat{\mathit{x}}(t) ; \omega(t)))$
		\State Update the generator by descending the stochastic gradient:
		\State $\nabla_{\psi} \log (1-D(\hat{\mathit{x}}(t) ; \omega(t))$
		\EndFor
			
	\end{algorithmic}
\end{algorithm}

\subsection{Hierarchical Structure and Anomaly Detection}\label{subsec: HGAN}
\begin{figure}
	\centering
	\includegraphics[width=\linewidth]{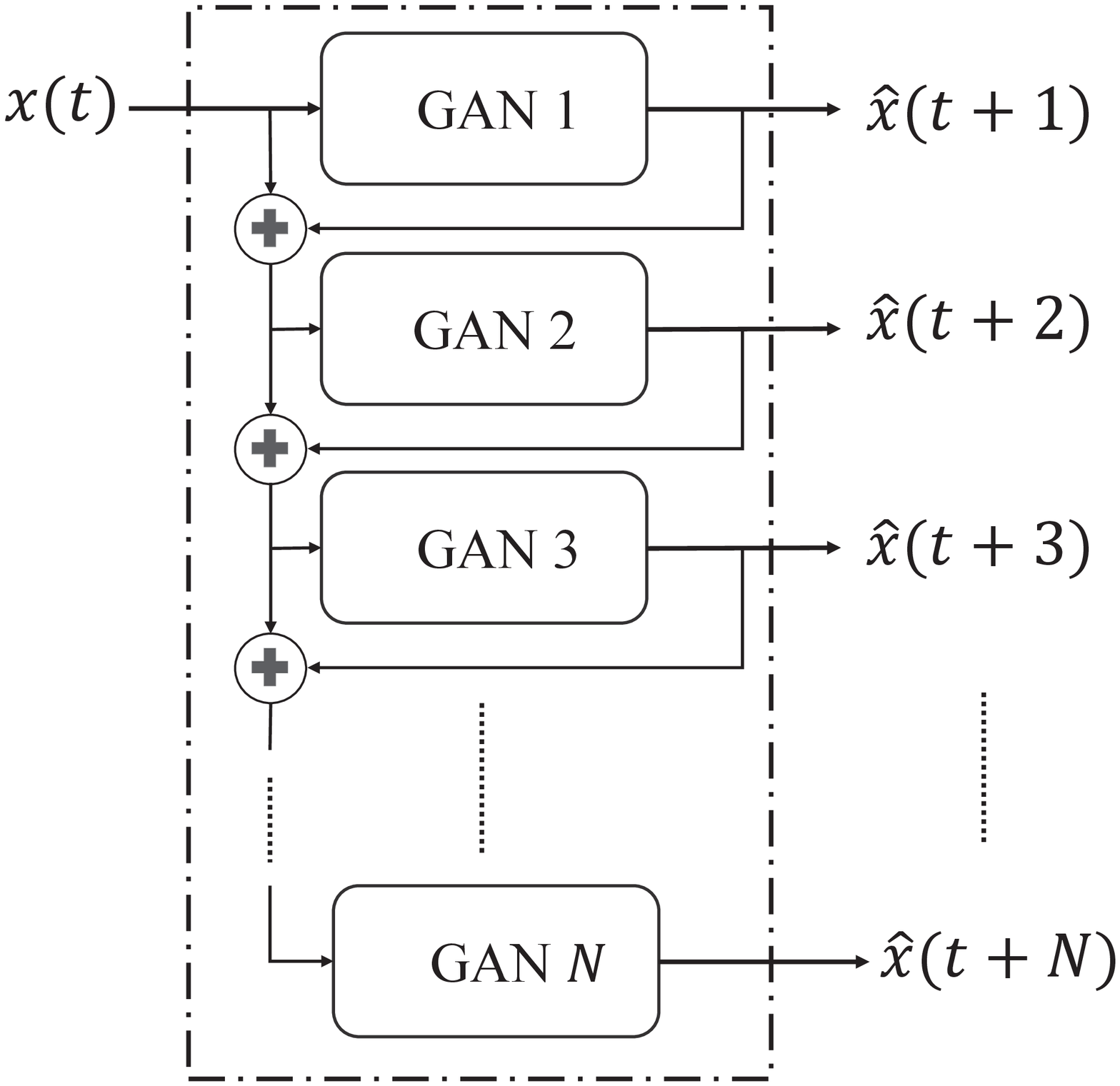}
	\caption{Structure of hierarchical GAN.}
	\label{fig:hgan}
\end{figure}

With the GAN introduced above, the sample $\mathit{x}(t)$ collected at time $t$ can be used to generate the predictions of data in the next $N$ time instants. Therefore, the choice of the value of $N$ determines the maximum gap between the two successive sampling time instants. To control the sampling cost in the anomaly detection, we assume that the system only takes one sample in each sampling time. Note that we can increase the prediction length $N$ to enable the detector to potentially sample less frequently. However, since a single sample contains very limited information for the GAN to make predictions, a single GAN will not be able to maintain a high prediction accuracy with increased prediction length. To achieve a better balance in this trade-off, we propose a hierarchical GAN structure as shown in Fig. \ref{fig:hgan}.


The hierarchical GAN consists of $N$ GANs, where GAN $i$ takes the sample collected at time $t$ and the predictions from the lower level GANs $1$ through $i-1$ as the input and generates the next prediction $\hat{\mathit{x}}(t + i)$. In this way, the hierarchical GAN takes advantage of the accurate predictions generated by the lower level GANs to reconstruct the pattern of the data and improve the reliability of the predictions. Meanwhile, to minimize the loss presented in (\ref{eq:sumLoss}), we add a squared error term to the loss function given in (\ref{eq:l_g}), so the loss of the generator in GAN $i$ is given as
\begin{equation}
\mathcal{L}_{G_i} =  \log (1-D(\hat{\mathit{x}}(t + i))) + (\hat{\mathit{x}}(t + i) - \mathit{x}(t + i))^2.
\end{equation}

While accurate predictions of the lower level GANs can help to reduce the prediction losses of the upper level GANs, if the lower level GANs are not well trained, the errors will propagate to the upper levels which can eventually lead to a large accumulated error at the last level. To address this, we train the hierarchical GAN level by level, and freeze the update of well-trained levels to avoid overfitting. The training procedure of hierarchical GAN is shown in Algorithm \ref{alg:TrainHGAN}. In the training phase, the OU processes are generated with random parameters $\sigma$ and $\theta$ at the beginning of every episode and are assumed to be available to the system.

\begin{algorithm}
	\caption{Training procedure of hierarchical GAN}
	\label{alg:TrainHGAN}
	\begin{algorithmic}
		\For{$i = 1 : N$}
		\State Initialize the generator $G_i(\mathit{s}; \psi_i)$ with the parameters $\psi_i$, and the the discriminator $D_i(\mathit{x};\omega_i)$, parameterized by $\omega_i$.
		\For{$\tau = 1$:Maximum episode}
		\State Randomly generate an OU process time series with length as $L$.
		\State Take $\mathit{x}(0)$ as the first sample and set $t = 0$.
		\While{$t \leq (L - i - 1)$}
		\For{$j = 1:i$}
		\State Fetch the input
		\State \qquad $\mathit{s}_j = [\mathit{x}(t), \hat{\mathit{x}}(t + 1), \dots, \hat{\mathit{x}}(t + j - 1)]$.
		\State Obtain $\hat{\mathit{x}}(t+j) =  G_j(\mathit{s}_j; \psi_j)$.
		\If{$j == i$}
		\State Update GAN $j$.
		\EndIf
		
		\EndFor
		\State Determine the next sampling time $T(\mathit{x}(t))$ using Eq. (\ref{eq:nextTime}).
		\State Set $t = T(\mathit{x}(t))$.
		
		\EndWhile
		
		\EndFor
		
		\EndFor
				
	\end{algorithmic}
\end{algorithm}

After the GANs are trained, the parameters are saved for the testing phase. In the testing phase, the OU processes are also generated with random parameters in every episode, but the parameters as well as the real data are no longer available to the hierarchical GAN. Unlike in the training phase, only the GANs that can make use of the available real samples can update the neural networks. For example, with real data $\mathit{x}(t)$, the system can determine the next sampling time $T(\mathit{x}(t))$. If we denote the next sample as $\mathit{x}(T)$, then the GAN which predicts the corresponding $\hat{\mathit{x}}(T)$ can use this projection and real data pair to update its model.

\section{Simulation Results}
\subsection{Experiment Settings}
\subsubsection{Environment}
In the experiments, we let the mean value $\mu$ be fixed at $0$, set the range of $\theta$ as $[0.02, 0.03]$, and the range of $\sigma$ as $[0.4, 0.6]$. At the beginning of each episode, the OU process will be generated with $\theta$ and $\sigma$ randomly selected from their corresponding range according to a uniform distribution. We also set the threshold as $\Gamma = 0$ throughout the experiments.

\subsubsection{Structure of GAN}

Each GAN consists of two neural networks: generator network and discriminator network. The input size of the generator network depends on the index of its level in the hierarchical structure. And in the generator network, there is one long short-term memory (LSTM) layer followed by two fully-connected layers. And the discriminator network consists of three fully-connected layers, and the ReLU activation function is employed in between layers and Sigmoid activation function is adopted after the output layer.

\subsection{Training Phase}
\begin{figure}[t]
	\centering
	\includegraphics[width=.8\linewidth]{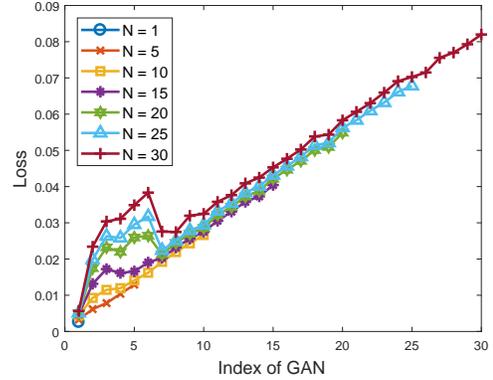}
	\caption{The prediction loss in each level of the hierarchical GAN.}
	\label{fig:loss}
\end{figure}

In the experiments, we first train each GAN in the hierarchy for $5000$ times, and when all GANs are converged, we freeze the update of GANs but continue feeding the OU process data to the detector. Then we record the squared error $(\hat{\mathit{x}}(t + i) - \mathit{x}(t + i))^2$ from each GAN as the loss. In Fig. \ref{fig:loss}, we plot the loss as a function of the GAN index, and compare the impact of increasing the prediction length $N$. It is obvious that as the index of the GAN increases, the loss tends to get accumulated. We also notice that when the index is between $2$ to $6$, the losses first increase and then drop to a lower level, and following this, the losses continue increasing at a fixed rate.

As we mentioned before, a single sample can only provide limited information for the GANs to predict the future samples, and consequently the loss jumps to relatively high levels initially. However, as more predictions are used as input to the upper level GANs, the loss is corrected to some degree by the LSTM layers. This is because of the fact that even though the previous prediction is not perfect, the presence of such prediction can act as a projection of the real data set to provide the upper level GANs with more features of the time series data.

\subsection{Testing Phase}
Considering the losses shown in Fig. \ref{fig:loss}, it is expected that the proposed GAN-based detector will experience errors in the testing phase. To reduce such errors, we introduce a small buffer zone of width $\rho$ around the threshold $\Gamma$ in the following experiments, and define the threshold-crossing time as the first time instant at which the predicted sample is within the range $[\Gamma - \rho, \Gamma + \rho]$. With this, the delay in detecting a threshold crossing is reduced at the cost of increased number of samples.

Random fluctuations of the stochastic process imply that the process can potentially cross over the threshold multiple times within a certain time frame, and duration from one threshold crossover to another varies randomly as well. Therefore, there are two potential outcomes of detection: 1) the GAN-based detector successfully detects the threshold crossing potentially with a delay but before another crossover occurs; and 2) the detector fails to detect the threshold crossing before another crossing occurs (and we indicate this outcome as \emph{failed/missed detection}).

\begin{figure}[t]
	\centering
	\includegraphics[width=.8\linewidth]{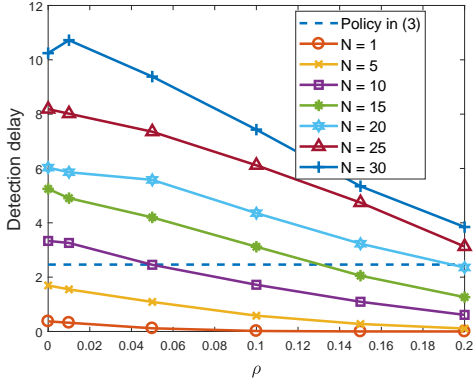}
	\caption{The average detection delay vs. buffer zone width $\rho$.}
	\label{fig:delay}
\end{figure}

We define the detection delay as the difference between the time instant when the change is detected and the time instant the change actually occurs. Thus, the delay varies between 0 (indicating perfect detection) and the time until a new crossing occurs (indicating that detection was not done before a new crossing). In Fig. \ref{fig:delay}, we plot the average detection delay as a function of the buffer zone width $\rho$ for different values of $N$. As we increase $\rho$, the detection delay achieved by the proposed hierarchical GAN-based detector decreases in all cases, and a delay smaller than that achieved by the sampling policy in (\ref{eq:approx_solution}) can be attained when the prediction length (or equivalently the number of GANs in the hierarchy) is $N = 1, 5, 10, 15$ or $20$ for sufficiently large $\rho$. Note that the sampling policy in (\ref{eq:approx_solution}) assumes complete statistical knowledge (which is not available to the GAN-based detector) but does not perform any explicit predictions.
In Fig.  \ref{fig:delay}, we further  observe that for fixed $\rho$, delays expectedly grow as we increase the prediction length $N$ and take fewer samples.

\begin{figure}[t]
	\centering
	\includegraphics[width=.8\linewidth]{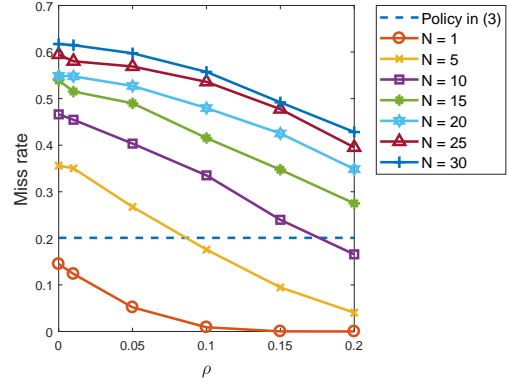}
	\caption{The average miss detection ratio vs. buffer zone width $\rho$.}
	\label{fig:miss}
\end{figure}

Since the detection delay is only considered when the threshold crossings are successfully detected, in Fig. \ref{fig:miss} we plot the miss rate to have a better understanding on the failed/missed detection rates. The miss rate is defined as the ratio of the number of crossings that are missed by the detector to the total number of crossings. We observe that the miss rates achieved with different values of prediction length $N$ decrease as the buffer size $\rho$ increases, and the miss rates for $N = 1, 5, 10$ can be lower than that achieved by the sampling  policy in (\ref{eq:approx_solution}). To further reduce the miss rate, we can continue increasing the buffer size, but this will lead to high sampling rates. We also notice in Fig. \ref{fig:miss} that miss rate increases as $N$ is increased. As noted above, with larger prediction length, the hierarchical GAN can sample less frequently. However, this increases the risks of miss detection because the duration until a new crossing can be far smaller than the prediction length, and when the short duration is coupled with the low sampling ratio, the changes are missed with an increased probability.

\begin{figure}[t]
	\centering
	\includegraphics[width=.8\linewidth]{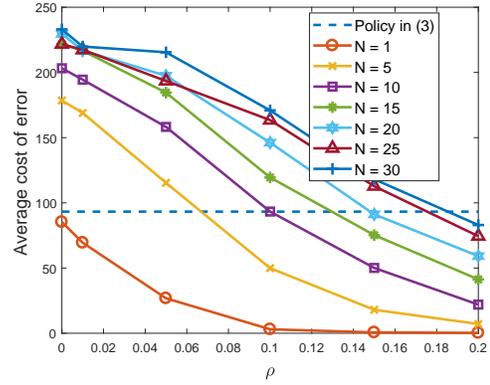}
	\caption{The average cost of error (due to delayed detection) vs. buffer zone width $\rho$.}
	\label{fig:cost}
\end{figure}

We can also measure the performance of the proposed GAN-based detector by considering the cost of error (due to delayed detection) as formulated in (\ref{eq:cost}).  Note that even if the miss rates are high with large prediction length, misses might occur due to short durations between consecutive threshold crossings, whose cost with respect to the metric in (\ref{eq:cost}) is small. To address this possibility and understand the impact of miss rates, we investigate the cost of error due to delayed detection. In Fig. \ref{fig:cost}, average costs are plotted as a function of $\rho$ for different values of $N$. Here, cost of error in (\ref{eq:cost}) is averaged over $10000$ time series. As seen in Fig. \ref{fig:cost}, the GAN-based detectors' performance in terms of costs approaches and exceeds the performance of the policy in (\ref{eq:approx_solution}) (i.e., starts achieving lower cost) as the buffer zone width increases. With buffer width set as $\rho = 0.1$, three out of the seven tested GAN-based detectors can perform competitively or better in comparison with the sampling policy in (\ref{eq:approx_solution}). The number increases to five when $\rho = 0.15$, and all seven GAN-based detector can work less costly with $\rho = 0.2$. On the other hand, in Fig. \ref{fig:miss}, less than half of the tested GAN-based detectors are able to outperform the policy in (\ref{eq:approx_solution}). This confirms that the number of missed detections are primarily due to short durations between consecutive threshold crossings.

\begin{figure}[t]
	\centering
	\includegraphics[width=0.8\linewidth]{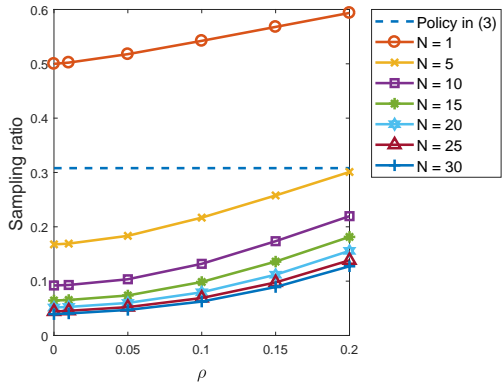}
	\caption{The average sampling ratio vs. buffer zone width $\rho$.}
	\label{fig:sampling}
\end{figure}

In the numerical analysis above, we have primarily addressed the performances in terms of detection delays. Note that prediction lengths also affect the sampling rates, which we investigate next numerically. In particular, we define the sampling ratio as the number of samples taken by the detector over the total number of samples in the time series. In Fig. \ref{fig:sampling}, we plot the sampling ratio required by the GAN-based detectors as a function of $\rho$. We observe that the sampling ratio grows with increasing $\rho$. We also see that except for the GAN-based detector with $N=1$, all other GAN-based detectors sample the data less frequently than policy in (\ref{eq:approx_solution}) for all values of $\rho$. Even after the sampling ratios grow with $\rho$ increasing to $0.2$, most of the GAN-based detectors still exhibit obvious advantages.

We have seen above that the GAN-based detectors' performance in terms of considered metrics are strongly influenced by the selection of $\rho$, and this makes $\rho$ a critical parameter enabling us to control the tradeoff between the delay costs and sampling rates. Specifically, we have observed in Figs. \ref{fig:delay}, \ref{fig:miss} and \ref{fig:cost} that the detection delays, miss rates, and average cost of error can all be reduced by increasing $\rho$ but at the expense of requiring more samples as seen in Fig. \ref{fig:sampling}. We have also noted that even though the GAN-based detectors do not have statistical knowledge of the OU processes, they can outperform the sampling policy in (\ref{eq:approx_solution}) in terms of detection delays and sampling ratio, owing to their well-trained neural networks and prediction capabilities.


\section{Conclusion}

In this paper, we have proposed a framework for anomaly detection and sampling cost control based on GANs. First, we have modeled the detection of threshold crossing in a stochastic time series as an anomaly detection problem. Then, we have proposed a hierarchical GAN structure to address such a detection problem. Specifically, we have designed a hierarchical structure to take advantage of the estimated projection of real samples and described the training and testing workflows. The performance of the proposed hierarchical GAN-based detector has been analyzed considering multiple performance metrics, namely the detection delay, miss rate, cost of error (due delayed detection) and sampling ratio. We have also provided comparisons between the proposed hierarchical GAN detector and the sampling policy derived with complete statistical information of OU processes. We have shown that the proposed GAN-based detector can have improved performance in terms of detection delays, miss rates, and cost of error as the buffer zone width is increased but at the cost of higher sampling ratios.

\bibliographystyle{ieeetr}
\bibliography{reference}

\end{document}